\documentclass[times,twocolumn,final]{elsarticle}

\usepackage{arxiv}
\usepackage{graphicx}%
\usepackage{multirow}%
\usepackage{amsmath,amssymb,amsfonts}%
\usepackage{amsthm}%
\usepackage{mathrsfs}%
\usepackage[title]{appendix}%
\usepackage{xcolor}%
\usepackage{textcomp}%
\usepackage{manyfoot}%
\usepackage{booktabs}%
\usepackage{algorithm}%
\usepackage{algorithmicx}%
\usepackage{algpseudocode}%
\usepackage{listings}%
\usepackage{multirow}
\usepackage{pifont}%
\usepackage{adjustbox}
\usepackage{graphicx}
\usepackage{hyperref}
\usepackage{float}
\usepackage{placeins}
\usepackage{fancyhdr}
\usepackage{dsfont}
\usepackage{threeparttable}
\usepackage{color,colortbl}
\definecolor{LightCyan}{rgb}{0.88,1,1}

\begin{document}
\fancypagestyle{firstpagestyle}{
    \fancyhf{} 
    \renewcommand{\headrulewidth}{0pt} 
    \fancyhead{} 
    \fancyfoot{} 
    \fancyhead[CO]{\em \fontsize{9pt}{8pt}\selectfont}
}

\fancypagestyle{default}{
    \fancyhf{}
    \fancyhead[R]{\thepage} 
    \renewcommand{\headrulewidth}{0.4pt} 
    \fancyhead[LO]{\em \fontsize{9pt}{8pt}\selectfont}
}



\title{fine-CLIP: Enhancing Zero-Shot Fine-Grained Surgical Action Recognition with Vision-Language Models}
\author[1,2]{Saurav \snm{Sharma} \fnref{corresp}}
\fntext[corresp]{Corresponding author: \texttt{ssharma@unistra.fr}}
\author[2,3]{Didier \snm{Mutter}}
\author[1,2]{Nicolas \snm{Padoy}}

\address[1]{University of Strasbourg, CNRS, INSERM, ICube, UMR7357, France}
\address[2]{IHU Strasbourg, Strasbourg, France}
\address[3]{University Hospital of Strasbourg, France}

\received{XXX}
\finalform{XXX}
\accepted{XXX}
\availableonline{XXX}
\communicated{XXX}

\begin{abstract}
While vision-language models like CLIP have advanced zero-shot surgical phase recognition, they struggle with fine-grained surgical activities, especially action triplets. This limitation arises because current CLIP formulations rely on global image features, which overlook the fine-grained semantics and contextual details crucial for complex tasks like zero-shot triplet recognition. Furthermore, these models do not explore the hierarchical structure inherent in triplets, reducing their ability to generalize to novel triplets.
To address these challenges, we propose fine-CLIP, which learns object-centric features and leverages the hierarchy in triplet formulation. Our approach integrates three components: hierarchical prompt modeling to capture shared semantics, LoRA-based vision backbone adaptation for enhanced feature extraction, and a graph-based condensation strategy that groups similar patch features into meaningful object clusters. Since triplet classification is a challenging task, we introduce an alternative yet meaningful base-to-novel generalization benchmark with two settings on the CholecT50 dataset: Unseen-Target, assessing adaptability to triplets with novel anatomical structures, and Unseen-Instrument-Verb, where models need to generalize to novel instrument-verb interactions. fine-CLIP shows significant improvements in F1 and mAP, enhancing zero-shot recognition of novel surgical triplets.
\\

\end{abstract}

\maketitle
\thispagestyle{firstpagestyle}

\section{Introduction}\label{intro}

Analyzing surgical workflows is vital in surgical data science~\cite{vercauteren2019cai4cai}, facilitating the development of context-aware decision systems that can aid surgeons in the operating room (OR). These systems utilize streaming video data to create comprehensive solutions for enhanced surgical decision-making. They focus on identifying surgical activities at various granularities, such as phases~\cite{twinanda2016endonet}, steps~\cite{Lavanchy2024}, and surgical action triplets~\cite{nwoye2021rendezvous,sharma2022rendezvous,sharma2023surgical}.
Vision-language (VL) models, such as CLIP~\cite{radford2021learning}, represent a significant shift from traditional vision-only models, enabling robust zero-shot image classification. This capability is especially valuable in the context of surgical procedures, where it allows for data-efficient learning despite limited labeled data and high annotation costs. By leveraging VL models, adaptable systems can interpret complex surgical triplets without the need for extensive retraining. These models are trained on large image-text datasets using contrastive learning, enabling them to encode open-vocabulary concepts in a shared semantic space. Building on this foundation, 
SurgVLP~\cite{yuan2023learning}, trained on surgical image-text data, surpasses CLIP in surgical phase and triplet recognition. Subsequent works~\cite{yuan2024hecvl,yuan2024procedure,hu2024ophclip} enhanced performance using CLIP-style contrastive learning with ResNet-50 and BioClinicalBERT encoders, representing various hierarchical levels in surgical videos and improving zero-shot phase recognition. VidLPRO~\cite{honarmand2024vidlpro} further advanced multi-modal representation by introducing multiple pretraining objectives alongside contrastive learning.

However, zero-shot performance on surgical action triplets~\cite{nwoye2021rendezvous} remains challenging. These triplets, defined as \textlangle{}{\textit{instrument, verb, target}}\textrangle{}, represent fine-grained tool-tissue interactions in surgical scenes, requiring models to disentangle and associate multiple concepts. Current approaches like CLIP and SurgVLP face two key limitations: (1) they rely on flattened features from ViT-B's~\cite{vit} CLS token, failing to capture fine-grained interactions, and (2) their zero-shot evaluation doesn't effectively test generalization. A more practical approach would be to fine-tune on a subset of triplets and evaluate on unseen ones, better reflecting real-world adaptability to new surgical interactions.

\begin{figure*}[h!]
    \centering
    \includegraphics[width=0.95\linewidth]{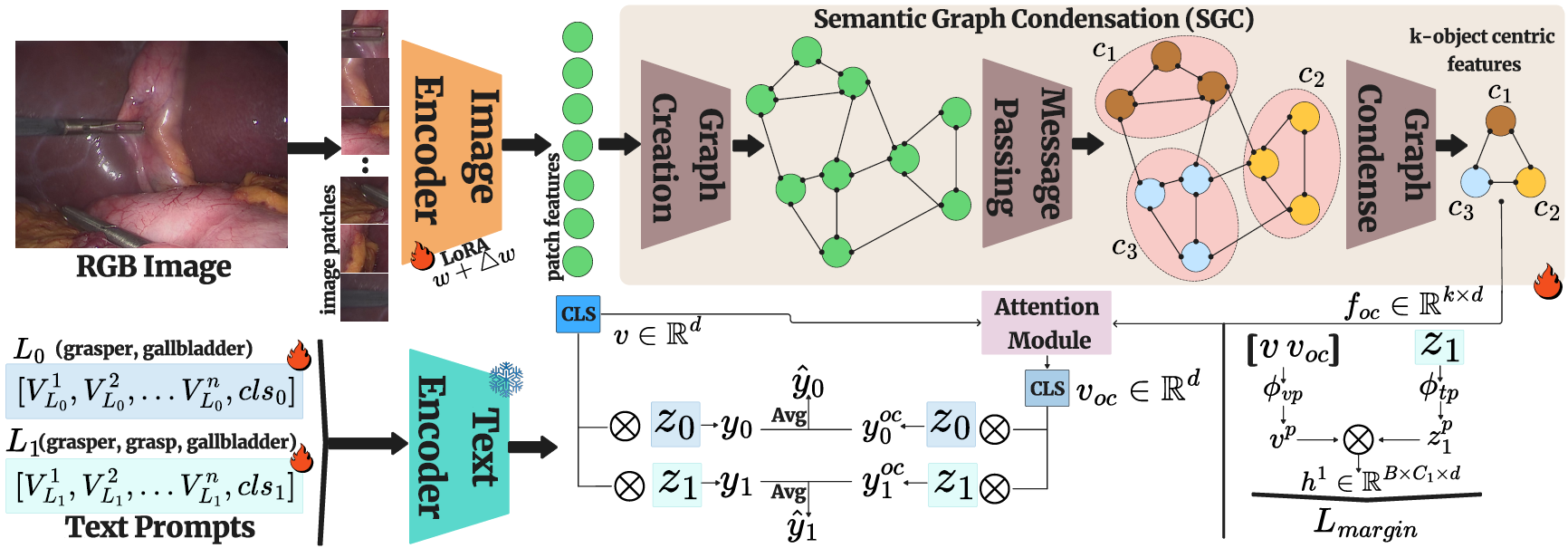}
    \caption{{\textbf{Model Overview:}} \textbf{fine-CLIP} tunes the vision backbone using LoRA~\cite{zanella2024low} and extracts object-centric features via Semantic Graph Condensation (SGC). Hierarchical prompts generate two-level text embeddings, while object features enhance image features through attention. A hierarchical margin loss on the combined object-aware and levelwise logits guides the final prediction.}
    \label{proposed_method}
\end{figure*}

Fine-tuning large-scale VL models is a challenging task, prompting recent research~\cite{zhou2022learning,zanella2024low} to develop efficient strategies that preserve CLIP's zero-shot capabilities while boosting performance on new tasks. These strategies are divided into two main categories: \textbf{prompt-learning} and \textbf{adaptation}. 
\textbf{Prompt-learning} methods, such as CoOp~\cite{zhou2022learning} and DualCoOp~\cite{sun2022dualcoop}, involve learning prompt vectors that integrate with class prompts, allowing models to adapt to downstream tasks without altering the visual and text encoders. MaPLe~\cite{khattak2023maple} adds multi-modal learnable prompts to both text and vision encoders in CLIP. \textbf{Adaptation} methods, like CLIP-Adapter~\cite{gao2024clip}, introduce a learnable projection to the vision encoder's features, keeping other weights frozen. LoRA~\cite{zanella2024low,hu2021loralowrankadaptationlarge} employs rank decomposition matrices within transformer layers for improved task adaptation. 
Despite these advancements, current methods for zero-shot triplet recognition often fall short, failing to capture the rich context of surgical scenes or utilize the hierarchical structure in triplet annotations. Addressing these limitations, we propose two research questions: (1) How can CLIP's patch features be used to develop object-centric features that enhance fine-grained detail recognition and improve zero-shot triplet performance? (2) Can we exploit the hierarchical relationships in triplet labels for better text and visual feature alignment?

To address these research questions, we introduce \textbf{fine-CLIP}, which adapts both text and vision backbones. It integrates learnable prompt vectors in the text encoder, similar to CoOp~\cite{zhou2022learning}, and employs LoRA~\cite{zanella2024low,hu2021loralowrankadaptationlarge} style low-rank decomposition in the vision encoder. We observe that patch features from CLIP's vision transformer attention modules contain rich object semantics~\cite{gandelsman2023interpreting}, enhancing global image features. Leveraging these, we develop \textbf{SGC} (Semantic Graph Condensation), which constructs a patch affinity graph, performs message passing, and applies graph condensation to cluster similar patch features into object-centric features. These features, combined with global image features through attention mechanisms, improve triplet understanding and boost zero-shot performance. Surgical action triplets exhibit inherent hierarchy; for example, \textlangle{}\textit{grasper, gallbladder}\textrangle{} includes multiple triplets with verbs like \textit{grasp} and \textit{retract}. We leverage this by designing learnable prompts at both \textlangle{}\textit{instrument, target}\textrangle{} (root) and \textlangle{}\textit{instrument, verb, target}\textrangle{} (leaf) levels, capturing broad interactions and specific actions. A max margin loss~\cite{vendrov2016orderembeddingsimageslanguage} aligns features with shared parents, enabling hierarchical knowledge transfer for zero-shot classification.

We evaluate fineCLIP's zero-shot generalization across two settings: \textbf{unseen target} (UT), where the model encounters new target tissues (e.g., trained on \textit{gallbladder}, tested on \textit{cystic-artery}), and \textbf{unseen instrument \& verb} (UIV), involving novel instrument-verb combinations. In both settings, our method outperforms existing approaches, including SurgVLP~\cite{yuan2023learning} and its variants.

\section{Methodology}
We introduce \textbf{fine-CLIP}, a zero-shot triplet classification model that builds upon CLIP~\cite{radford2021learning}. CLIP employs dual encoders: an image encoder $\Phi_{\text{img}}$ based on Vision Transformer (ViT)~\cite{vit} and a text encoder $\Phi_{\text{text}}$, each comprising $M$ transformer layers. For visual processing, CLIP divides an image $I \in \mathbb{R}^{C \times H \times W}$ into $N$ patches, adds a class token $p_{\text{cls}}$, and projects them to $\tilde{I} = \{p_{\text{cls}}, p_1, \ldots, p_N\}$, creating a visual embedding $v = \Phi_{\text{img}}(\tilde{I}) \in \mathbb{R}^d$. For text, CLIP encodes a label $y$ using the template "a photo of a \{class label\}" into $\tilde{T} = \{s_{\text{start}}, w_1, \ldots, w_L, l_y, s_{\text{end}}\}$, with $L$ as the prompt length, yielding a text embedding $t = \Phi_{\text{text}}(\tilde{T}) \in \mathbb{R}^d$.

\subsection{Hierarchical Prompt Learning:}
Recent work, such as CoOp~\cite{zhou2022learning}, shows that soft prompting can effectively adapt text encoders with minimal overhead. Given the complexity of surgical triplets, a hierarchical prompting approach captures multi-level relationships for zero-shot classification. This hierarchy includes level $L_1$ for fine-grained triplet interactions \textlangle{}{\textit{instrument, verb, target}}\textrangle{} and level $L_0$ for coarse-grained instrument-tissue relationships \textlangle{}{\textit{instrument, target}}\textrangle{}. The text encoder $\Phi_{\text{text}}$ processes $t_{0} = \{V^1_{L_0}, \dots, V^n_{L_0}, cls_{0}\}$ and $t_{1} = \{V^1_{L_1}, \dots, V^n_{L_1}, cls_{1}\}$ into embeddings $z_{0}, z_{1} \in \mathbb{R}^d$, where $cls_{0}$ and $cls_{1}$ are class embeddings for levels $L_0$ and $L_1$, respectively, with $n$ context tokens (Figure~\ref{proposed_method}).

\subsection{Low-Rank Adaptation:}
To efficiently adapt large-scale models without full finetuning, we employ Low-Rank Adaptation (LoRA)~\cite{hu2021loralowrankadaptationlarge}. LoRA approximates weight updates through low-rank decomposition using matrices \textbf{A} $\in \mathbb{R}^{r \times d}$ and \textbf{B} $\in \mathbb{R}^{d \times r}$, where rank $r \ll d$ captures essential parameter updates. For each weight matrix \textbf{W}, the update is applied as $\textbf{W} + \alpha\textbf{BA}$, with $\alpha$ as a scaling factor. Following CLIP-LoRA~\cite{zanella2024low}, we adapt the attention modules' query, key, and value matrices using low-rank decomposition with rank $r = 2$.

\subsection{Learning Object-Centric Features:} CLIP's ViT-B attention key features naturally encode object-level cues~\cite{gandelsman2023interpreting}. Building on this, our Semantic Graph Condensation (\textbf{SGC}) module learns object-centric features without instance supervision. SGC constructs an undirected graph $\mathcal{G}$ from patch features of the $j^{th}$ layer attention key using pairwise cosine similarities between patches. Node features are enriched through Graph Attention Networks (GAT)~\cite{velivckovic2017graph} to aggregate contextual semantics from neighboring patches. Following graph condensation~\cite{jin2021graph}, a softmax-normalized assignment matrix maps nodes to $k$ clusters, producing object-centric features $f_{oc} \in \mathbb{R}^{k \times d}$ through matrix multiplication.

\subsection{Hierarchical Feature Alignment:} Given object-centric features $f_{oc}$, visual embedding $v$, and hierarchical text embeddings $z_0$ and $z_1$, we compute logits in two stages. First, following CLIP's approach, we calculate image logits by finding the similarity between L2-normalized image features and text embeddings: $y_1 = \tau(v \cdot z_1^T)$ and $y_0 = \tau(v \cdot z_0^T)$, where $\tau$ is a scaling factor. We then enrich image features with object-centric information by computing attention weights: $\alpha = \text{softmax}\left(\frac{v \cdot f_{oc}^T}{\sqrt{d}}\right)$. These weights yield object-aware features $v_{oc} = \alpha f_{oc}$, used to compute another set of logits: $y^{oc}_1 = \tau(v_{oc} \cdot z_1^T)$ and $y^{oc}_0 = \tau(v_{oc} \cdot z_0^T)$. The final logits are averaged: $\hat{y}_i = \frac{1}{2}(y_i + y^{oc}_i)$ for $i \in \{0,1\}$.

\subsection{Optimization:}
We apply binary cross entropy loss between predicted logits $\hat{y}_i$ and ground truth labels $y_i$ at each level $i \in \{0,1\}$ (Equation~\ref{bce_formula}):
\begin{equation}
\label{bce_formula}
L_i = \sum_{c=1}^{C_i}\frac{-1}{N}\left(y_{i,c}\log\left(\sigma(\hat{y}_{i,c})\right) + (1-y_{i,c})\log\left(1-\sigma(\hat{y}_{i,c})\right)\right),
\end{equation}
where $C_i$ refers to the number of classes at level $i$, $y_{i,c}$ and $\hat{y}_{i,c}$ denote ground truth and predicted labels for class $c$ respectively, and $\sigma$ is the sigmoid function. 
The levels $L_0$ and $L_1$ exhibit a natural hierarchical relationship, where fine-grained triplets \textlangle{}{\textit{instrument, verb, target}}\textrangle{} are subsets of coarse-grained pairs \textlangle{}{\textit{instrument, target}}\textrangle{}. To leverage this hierarchy and enforce hierarchical constraints, we introduce a max-margin loss operating on level $L_1$ embeddings. For triplets sharing the same parent (e.g., triplets with same \textlangle{}{\textit{instrument, target}}\textrangle{} but different verbs), their embeddings should be closer compared to triplets with different parents. We concatenate image features $v$ with object-aware features $v_{oc}$ and project them to $\mathbb{R}^{\hat{d}}$ using $\phi_{vp}$ to obtain $v^{p}$. Similarly, text embeddings $z_1$ are projected through $\phi_{tp}$ to obtain $z^{p}_{1}$. For level $L_1$, we compute class embeddings $h^1 = v^p \cdot z^{p}_{1}$, $h^1 \in \mathbb{R}^{B \times C_1 \times d}$. The hierarchical loss (Equation~\ref{hierarchical_loss}) is defined as:
\begin{equation}
\label{hierarchical_loss}
L_{margin} = \frac{1}{|P|} \sum_{(i,j) \in P} \sum_{k \notin N_{ij}} \max(0, m + d(h^1_i, h^1_j) - d(h^1_i, h^1_k)),
\end{equation}
where $P$ is the set of positive pairs sharing the same parent within a batch, $N_{ij}$ contains batch-sampled negative triplets not sharing parents with $i$ or $j$, $d(\cdot,\cdot)$ is cosine distance, and $m$ is the margin.
The final loss combines hierarchical and classification objectives (Equation~\ref{final_loss}):

\begin{equation}
\label{final_loss}
L_{total} = \alpha_{0} \times L_0 + \alpha_{1} \times L_1 + \alpha_{h} \times L_{margin},
\end{equation}
where $\alpha_{0}$, $\alpha_{1}$, and $\alpha_{h}$ are the respective loss weights.

\section{Experiments}

\subsection{Base-to-Novel Generalization:}
We evaluate zero-shot generalization by splitting the 100 triplet classes into base and novel sets. The model is trained solely on the base set and then evaluated on the novel set without additional training.
In the \textbf{Unseen-Target} (UT) setting, we test generalization to new anatomical targets: the base set (36 triplets) includes interactions with \textit{gallbladder}, \textit{cystic duct}, \textit{cystic pedicle}, \textit{liver}, \textit{omentum}, and \textit{blood vessel}, while the novel set (18 triplets) contains \textit{abdominal wall cavity}, \textit{cystic artery}, \textit{cystic plate}, \textit{fluid}, \textit{gut}, \textit{peritoneum}, and \textit{specimen bag}.
In the \textbf{Unseen-Instrument-Verb} (UIV) setting, we evaluate generalization to new surgical interactions: the base set (28 triplets) includes instrument-verb pairs like \textit{bipolar-coagulate/dissect} and \textit{grasper-dissect/grasp}, while the novel set (21 triplets) features previously unseen combinations such as \textit{irrigator-aspirate/irrigate}, \textit{grasper-retract/pack}, and \textit{scissors-cut}.

\subsection{Dataset and Evaluation Metrics:}
We evaluate on the CholecT50~\cite{nwoye2021rendezvous} dataset following RDV splits~\cite{ctsplits}. To prevent data leakage, we exclude images containing both base and novel triplets. In the UT setting, we have $\sim$45K, $\sim$3.5K, and $\sim$12K frames for base training, validation, and test sets, respectively, plus $\sim$0.7K and $\sim$1.9K frames for novel validation and test sets. In the UIV setting, we obtain $\sim$18K, $\sim$1.7K, and $\sim$4.3K frames for base training, validation, and test sets, along with $\sim$1K and $\sim$3.7K frames for novel validation and test sets. We measure performance using mean average precision (mAP) and F1-Score@3, and report base, novel, and their harmonic mean (HM) scores on the test set.

\subsection{Implementation Details:}
We implement fine-CLIP in PyTorch and train on a single NVIDIA-A100 GPU. We resize input frames to $224 \times 224$ and apply RandAugment. Using ViT-B/16 as the image encoder, we train for 30 epochs with AdamW (learning rate: $1e^{-3}$, weight decay: $1e^{-2}$) optimizer and a batch size of 32. The text encoder is the default Transformer used in CLIP~\cite{radford2021learning}. We use $n=8$ context tokens and set a margin $m=0.7$ in $L_{margin}$. We set $d=512$ and $\tilde{d}=256$, using 1-layer MLPs for $\phi_{vp}$ and $\phi_{tp}$. SGC uses one GAT~\cite{velivckovic2017graph} layer for message passing and condensation. We tune hyperparameters on the validation set, setting all loss weights ($\alpha_{0}$, $\alpha_{1}$, $\alpha_{h}$) to 1, while baselines use default settings.

\begin{table*}[t]
    \centering
    \setlength{\tabcolsep}{5pt}
    \caption{\textbf{Zero-shot triplet classification performance (F1@3 and AP in \% ) on UT setting (P: prompt learning, A: adaptation, ZS: zero-shot, FT: Fine-tuning).}}
    \label{ut_result}
    \resizebox{\textwidth}{!}{%
    \begin{tabular}{@{}lccrccrccrcc@{}}
        \toprule
        \multirow{2}{*}{Method} &
        \multirow{2}{*}{\parbox{2cm}{\centering \#Trainable Params(MB)}} & 
        \multirow{2}{*}{\parbox{2cm}{\centering Category}} & 
        \phantom{abc} & \multicolumn{2}{c}{Base} &
        \phantom{abc} & \multicolumn{2}{c}{Novel} &
        \phantom{abc} & \multicolumn{2}{c}{HM}\\
         \cmidrule{5-6}\cmidrule{8-9}\cmidrule{11-12} 
             &&&& $F1@3$ & $mAP$ && $F1@3$ & $mAP$ && $F1@3$ & $mAP$ \\ \midrule
             CLIP~\cite{radford2021learning} & 0 & ZS && 0.73 & 5.64 && 9.2 & 5.85 && 1.34 & 5.74\\
             CLIP~\cite{radford2021learning} & 86.192 & FT && 57.36 & 19.29 && 24.12 & 18.05 && 33.96 & 18.64\\
             CoOp~\cite{zhou2022learning} & 0.004 & P  && 50.89 & 8.52 && 22.88 & 10.35 && 31.56 & 9.35\\
             DualCoOp~\cite{sun2022dualcoop} & 0.065 & P && 51.26 & 7.93 && 17.27 & 8.55 && 25.83 & 8.23\\
             MaPLe~\cite{khattak2023maple} & 3.555 & P && 58.89 & 21.85 && 11.35 & 18.23 && 19.03 & 19.87\\
             CLIP-Adapter~\cite{gao2024clip} & 0.131 & A && 0.32 & 5.76 && 19.62 & 8.00 && 0.62 & 6.70\\
             CLIP-LoRA~\cite{zanella2024low} & 0.184 & A && 57.04 & 18.04 && 16.13 & 19.22 && 25.15 & 18.61\\
             SurgVLP~\cite{yuan2023learning} & 0 & ZS && 2.38 & 5.81 && 8.43 & 7.45 && 3.71 & 6.53\\
             HecVLP~\cite{yuan2024hecvl} & 0 & ZS && 0.92 & 5.82 && 4.78 & 7.48 && 1.55 & 6.54\\
             PeskaVLP~\cite{yuan2024procedure} & 0 & ZS && 3.97 & 6.11 && 16.75 & 8.98 && 6.42 & 7.28\\
             SurgVLP~\cite{yuan2023learning} & 133.39 & FT && 54.75 & 13.68 && 17.73 & 17.46 && 26.79 & 15.34\\
             HecVLP~\cite{yuan2024hecvl} & 133.39 & FT && 55.11 & 12.53 && 26.55 & 14.28 && 35.83 & 13.35\\
             PeskaVLP~\cite{yuan2024procedure} & 133.39 & FT && 56.99 & 16.72 && 30.35 & 16.17 && 39.60 & 16.44\\
             \rowcolor{LightCyan} fine-CLIP (Ours) & 3.019 & PA && \textbf{61.71} & \textbf{31.72} && \textbf{39.78} & \textbf{32.17} && \textbf{48.38} & \textbf{31.95} \\
            \bottomrule
    \end{tabular}
    }
\end{table*}

\subsection{Results and Discussion}

\noindent\textbf{Baselines:}
We prepare a comprehensive set of baselines, including prompt learning methods like CoOp~\cite{zhou2022learning}, DualCoOp~\cite{sun2022dualcoop}, and MaPLe~\cite{khattak2023maple}, along with adaptation strategies such as CLIP-Adapter~\cite{gao2024clip} and CLIP-LoRA~\cite{zanella2024low}.
We also include SurgVLP~\cite{yuan2023learning} and its variants that have been trained on a large corpus of surgical image and text pairs and have shown improved zero-shot phase recognition. We finetune CLIP (CLIP-FT) and SurgVLP variants (denoted by FT) on the images of base triplets for comprehensive evaluation. 

\noindent\textbf{Unseen-Target (UT):} As shown in Table~\ref{ut_result}, fine-CLIP with only $3.019$M trainable parameters surpasses PeskaVLP (the best SurgVLP variant) by $+9.43$ percentage points (pp) in novel class F1@3, while showing even larger gains over prompt learning ($+16.90$ pp vs CoOp) and adaptation strategies ($+23.65$ pp vs CLIP-LoRA). This superior generalization yields $+8.78$ pp gain in harmonic mean F1@3 over PeskaVLP.

\noindent\textbf{Unseen-Instrument-Verb (UIV):} Table~\ref{uv_result} presents results for this challenging setting where models must generalize to unseen instrument-verb combinations. fine-CLIP demonstrates strong generalization to novel combinations, surpassing HecVLP by $+2.48$ pp in F1@3, while showing substantial gains over prompt learning ($+17.74$ pp vs MaPLe) and adaptation strategies ($+16.10$ pp vs CLIP-Adapter). This superior performance on novel combinations is achieved while preserving the best base performance, resulting in a $+2.95$ pp gain in harmonic mean F1@3 over HecVLP.
Across both settings, fine-CLIP demonstrates impressive zero-shot generalization, outperforming both lightweight adaptation methods (CoOp, LoRA) and heavily pretrained models (SurgVLP variants). This performance highlights the value of learning object-centric features and jointly tuning text and vision encoders for generalizing to novel surgical triplets.

\begin{table*}[t]
    \centering
    \setlength{\tabcolsep}{5pt}
    \caption{\textbf{Zero-shot triplet classification performance (F1@3 and AP in \% ) on UIV setting. (P: prompt learning, A: adaptation, ZS: zero-shot, FT: Fine-tuning).}}
    \label{uv_result}
    \resizebox{\textwidth}{!}{%
    \begin{tabular}{@{}lccrccrccrcc@{}}
        \toprule
        \multirow{2}{*}{Method} &
        \multirow{2}{*}{\parbox{2cm}{\centering \#Trainable Params(MB)}} &
        \multirow{2}{*}{\parbox{2cm}{\centering Category}} &
        \phantom{abc} & \multicolumn{2}{c}{Base} &
        \phantom{abc} & \multicolumn{2}{c}{Novel} &
        \phantom{abc} & \multicolumn{2}{c}{HM}\\
         \cmidrule{5-6}\cmidrule{8-9}\cmidrule{11-12} 
             &&&& $F1@3$ & $mAP$ && $F1@3$ & $mAP$ && $F1@3$ & $mAP$ \\ \midrule
             CLIP~\cite{radford2021learning} & 0 & ZS && 0.22 & 4.88 && 0.18 & 6.31 && 0.2 & 5.5\\
             CLIP~\cite{radford2021learning} & 86.192 & FT && 39.84 & 16.33 && 2.12 & 7.55 && 4.03 & 10.32\\
             CoOp~\cite{zhou2022learning} & 0.004 & P  && 37.91	& 11.94 && 1.94	& 7.77 && 3.69 & 9.42\\
             DualCoOp~\cite{sun2022dualcoop} & 0.065 & P && 37.81 & 12.41 && 2.06 & 6.8 && 3.91 & 8.79\\
             MaPLe~\cite{khattak2023maple} & 3.555 & P && 44.15	& 21.19 && 5.7 & 8.19 && 10.09 & 11.81\\
             CLIP-Adapter~\cite{gao2024clip} & 0.131 & A && 3.16 & 6.47 && 7.34	& 6.98 && 4.42 & 6.72\\
             CLIP-LoRA~\cite{zanella2024low} & 0.184 & A && 43.41 & 20.61 && 5.05 & 7.31 && 9.05 & 10.79\\
             SurgVLP~\cite{yuan2023learning} & 0 & ZS && 2.50 & 5.65 && 1.95 & 6.24 && 2.19 & 5.93\\
             HecVLP~\cite{yuan2024hecvl} & 0 & ZS && 2.63 & 6.12 && 4.70 & 6.50 && 3.37 & 6.30\\
             PeskaVLP~\cite{yuan2024procedure} & 0 & ZS && 5.93 & 6.25 && 20.73 & 7.03 && 9.23 & 6.62\\
             SurgVLP~\cite{yuan2023learning} & 133.39 & FT && 41.21 & 13.83 && 13.62 & 8.05 && 20.48 & 10.18\\
             HecVLP~\cite{yuan2024hecvl} & 133.39 & FT && 41.21 & 13.8 && 20.96 & 7.63 && 27.79 & 9.83\\
             PeskaVLP~\cite{yuan2024procedure} & 133.39 & FT && 41.62 & 17.86 && 19.94 & 9.21 && 26.96	& 12.15\\
             \rowcolor{LightCyan}fine-CLIP (Ours) & 3.019 & PA && \textbf{44.66} & \textbf{27.33} && \textbf{23.44} & \textbf{10.98} && \textbf{30.74} & \textbf{15.66} \\
            \bottomrule
    \end{tabular}
    }
\end{table*}

\begin{figure*}[t!]
    \centering
    \includegraphics[width=0.90\linewidth]{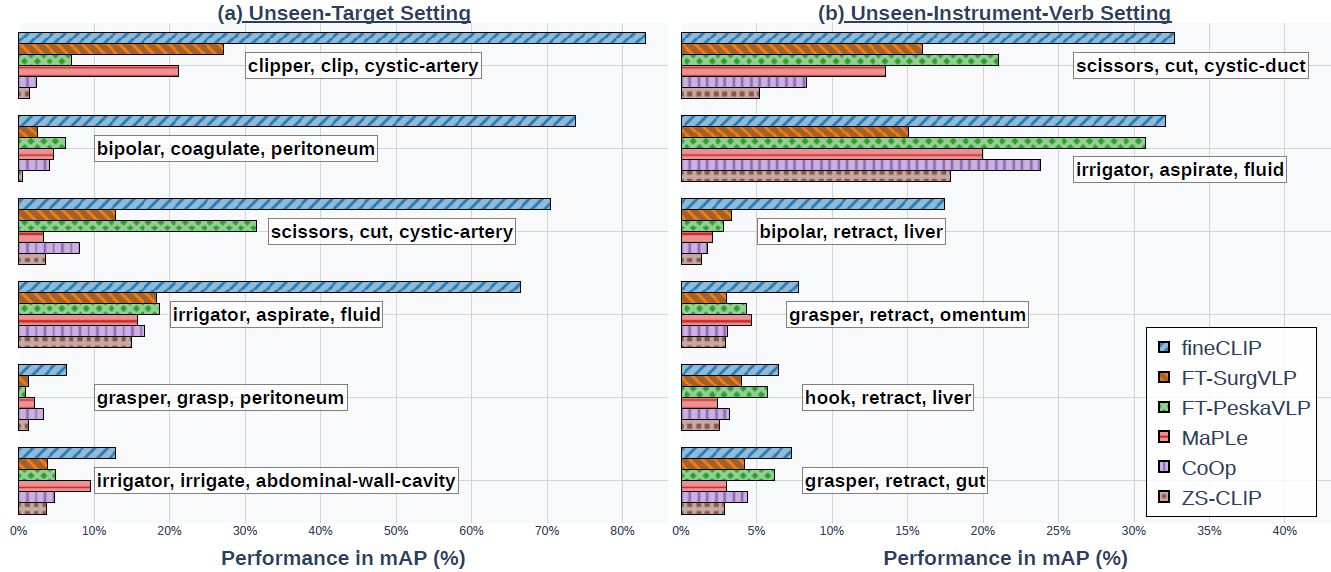}
    \caption{{Performance on novel triplets in (a) UT and (b) UIV settings.}}
    \label{classwise_performance}
\end{figure*}

\begin{figure*}[h]
    \centering
    \includegraphics[width=0.85\linewidth]{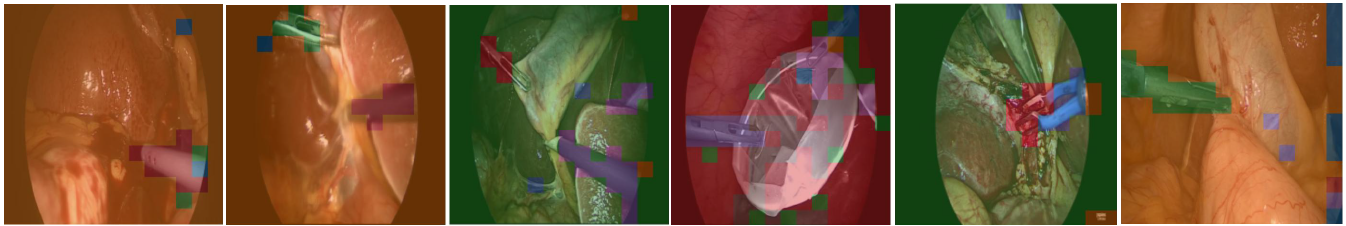}
    \caption{\textbf{Qualitative Results}: Visualization of the clusters. (Best viewed in color)}
    \label{qual_results}
\end{figure*}

\noindent\textbf{Analysis of Zero-shot generalization:}
As observed from Figure~\ref{classwise_performance}, fine-CLIP exhibits distinct generalization patterns across settings. In UT setting (a), it excels at precise surgical actions on anatomically related structures, with \textlangle{}\textit{clipper, clip, cystic-artery}\textrangle{} and \textlangle{}\textit{scissors, cut, cystic-artery}\textrangle{} achieving high performance ($\sim$80\% and $\sim$70\% mAP), benefiting from similar actions on cystic duct during training. Strong generalization is also seen for \textlangle{}\textit{irrigator, aspirate, fluid}\textrangle{} ($\sim$65\% mAP), while struggling with targets requiring different interaction patterns in \textlangle{}\textit{grasper, grasp, peritoneum}\textrangle{} ($\sim$5\% mAP). In UIV setting (b), the model effectively transfers specialized actions like \textlangle{}\textit{scissors, cut, cystic-duct}\textrangle{} ($\sim$33\% mAP) and \textlangle{}\textit{irrigator, aspirate, fluid}\textrangle{} ($\sim$32\% mAP), but shows limited performance on \textit{retract} actions across different instruments (\textlangle{}\textit{bipolar, retract, liver}\textrangle{}, \textlangle{}\textit{grasper, retract, omentum}\textrangle{}, $\sim$5-17\% mAP). This suggests fine-CLIP's effectiveness depends on both anatomical similarities in UT and action complexity in UIV. Figure~\ref{qual_results} visualizes the clusters in the test set of novel triplets, illustrating how SGC module effectively aggregates patches associated with instrument-tissue interactions.

\subsection{Ablation Study:} Table~\ref{ablation}(a) shows the impact of fine-CLIP components, revealing that incorporating hierarchical loss and object-centric features via SGC enhances zero-shot triplet generalization. Table~\ref{ablation}(b) evaluates layer-wise performance with optimal results at the $8^{th}$ ViT-B/16 layer. Table~\ref{ablation}(c) highlights model results across different cluster counts $k$. 

\section{Conclusion}
In this work, we address the limitations of surgical vision-language models and CLIP's prompt learning and adaptation strategies for zero-shot triplet classification. Existing methods overlook crucial patch-level object semantics by relying on flattened image features. We introduce fine-CLIP, a novel model that overcomes these limitations in three key ways: it learns object-centric features from a semantic patch graph using a graph condensation strategy, employs hierarchical prompts to capture hierarchical relationships, and presents a new base-to-novel generalization benchmark for evaluation. This benchmark assesses zero-shot triplet classification performance when trained on a subset of triplet classes, showing improved results over strong baselines.

\section{Acknowledgements}
This work was supported by French state funds managed by the ANR within the National AI Chair program under Grant ANR-20-CHIA-0029-01 (Chair AI4ORSafety) and within the Investments for the future program under Grant ANR-10-IAHU-02 (IHU Strasbourg). It was granted access to the GPU resources managed by CAMMA/IHU Strasbourg.

\begin{table*}[h!]
\centering
\caption{\textbf{Ablation Studies on fine-CLIP (Harmonic Mean: F1@3 and AP in \%).}}
\label{ablation}
\setlength{\tabcolsep}{9pt}
\resizebox{\textwidth}{!}{%
\begin{tabular}{@{}l @{\hspace{1em}} c @{\hspace{1em}} c @{\hspace{1em}} c @{\hspace{1em}} c @{\hspace{1em}} c @{\hspace{1em}} c | cccc | cccc @{}}
\toprule
\multicolumn{7}{c|}{3(a): Component-wise Results} & \multicolumn{4}{c|}{3(b): Per-Layer Results} & \multicolumn{4}{c}{3(c): \#Clusters ($k$)} \\
\toprule
& SoftPrompt & LoRA & Hierarchy & SGC & F1@3 & mAP & & j & F1@3 & mAP && $k$ & F1@3 & mAP \\ \midrule
& \checkmark & & & & 31.56 & 9.31 && 7 & 43.66 & 27.38 && 5 & 47.02 & 28.50 \\
& & \checkmark & & & 25.15 & 18.61 && 8 & \textbf{48.38} & \textbf{31.95} && 10 & 43.88 & 26.08 \\
& \checkmark & \checkmark & & & 38.44 & 20.61 && 9 & 33.97 & 28.20 && 15 & \textbf{48.38} & \textbf{31.95} \\
& \checkmark & \checkmark & \checkmark && 39.16 & 23.53 && 10 & 44.57 & 28.12 && 20 & 41.62 & 27.62 \\
& \checkmark & \checkmark & \checkmark & \checkmark & \textbf{48.38} & \textbf{31.95} && 11 & 42.17 & 24.93 && 25 & 43.90 & 28.87 \\
\bottomrule
\end{tabular}
}
\end{table*}

\noindent{\bf Code availability} Source code will be provided at \url{https://github.com/CAMMA-public/fine-CLIP}.

\bibliographystyle{sn-basic}
\bibliography{sn-bibliography}

\end{document}